\documentclass[letterpaper, 10 pt, conference]{ieeeconf}  

\IEEEoverridecommandlockouts                              

\usepackage{amsmath}
\usepackage{adjustbox}
\usepackage{booktabs} 
\usepackage{amsfonts} 
\usepackage{algorithmic, algorithm} 
\usepackage{bbm}
\usepackage{breqn}
\usepackage{graphicx}
\usepackage{xcolor}
\usepackage{pifont}

\newcommand{\cmark}{\ding{51}}%
\newcommand{\xmark}{\ding{55}}%

\title{\LARGE \bf
GO-Flock: Goal-Oriented Flocking in 3D Unknown Environments with Depth Maps
}

\author{Yan Rui Tan$^{1}$, Wenqi Liu$^{1}$, Wai Lun Leong$^{1}$, John Guan Zhong Tan$^{1}$, Wayne Wen Huei Yong$^{1}$, \\Shaohui Foong$^{2}$, Fan Shi$^{1}$ and Rodney Swee Huat Teo$^{1}$
\thanks{$^{1}$Yan Rui Tan, Wenqi Liu, Wai Lun Leong, John Guan Zhong Tan, Wayne Wen Huei Yong, Fan Shi and Rodney Swee Huat Teo are with the National University of Singapore,
        {\tt\small yr.tan@u.nus.edu, \{tslliuw, william.leong, johntgz, wayneyong, fan.shi, tsltshr\}@nus.edu.sg}}%
\thanks{$^{2}$ Shaohui Foong is with the Singapore University of Technology and Design,
{\tt\small foongshaohui@sutd.edu.sg}}%
}

\begin{document}

\maketitle
\thispagestyle{empty}
\pagestyle{empty}

\begin{abstract}

Artificial Potential Field (APF) methods are widely used for reactive flocking control, but they often suffer from challenges such as deadlocks and local minima, especially in the presence of obstacles. Existing solutions to address these issues are typically passive, leading to slow and inefficient collective navigation. As a result, many APF approaches have only been validated in obstacle-free environments or simplified, pseudo-3D simulations. This paper presents \textbf{GO-Flock}, a hybrid flocking framework that integrates planning with reactive APF-based control. GO-Flock consists of an upstream \textit{Perception Module}, which processes depth maps to extract waypoints and virtual agents for obstacle avoidance, and a downstream \textit{Collective Navigation Module}, which applies a novel APF strategy to achieve effective flocking behavior in cluttered environments. We evaluate GO-Flock against passive APF-based approaches to demonstrate their respective merits, such as their flocking behavior and the ability to overcome local minima. Finally, we validate GO-Flock through obstacle-filled environment and also hardware-in-the-loop experiments where we successfully flocked a team of \textbf{nine drones}—\textbf{six physical and three virtual}— in a forest environment.

\end{abstract}

\section{INTRODUCTION}

Flocking behavior, commonly observed in nature, involves the collective and coordinated movement of groups, such as flocks of birds or schools of fish. This natural phenomenon provides significant advantages that can be leveraged in multi-agent robotic systems. By mimicking flocking behavior, robotic systems can achieve sophisticated collective navigation, resulting in several key benefits such as efficient task distribution among multiple agents and lower production costs compared to a single, more complex system. Additionally, maintaining close proximity between agents enhances communication reliability.

   \begin{figure}[thpb]
      \centering
      \includegraphics[width=8.6cm, height=6cm]{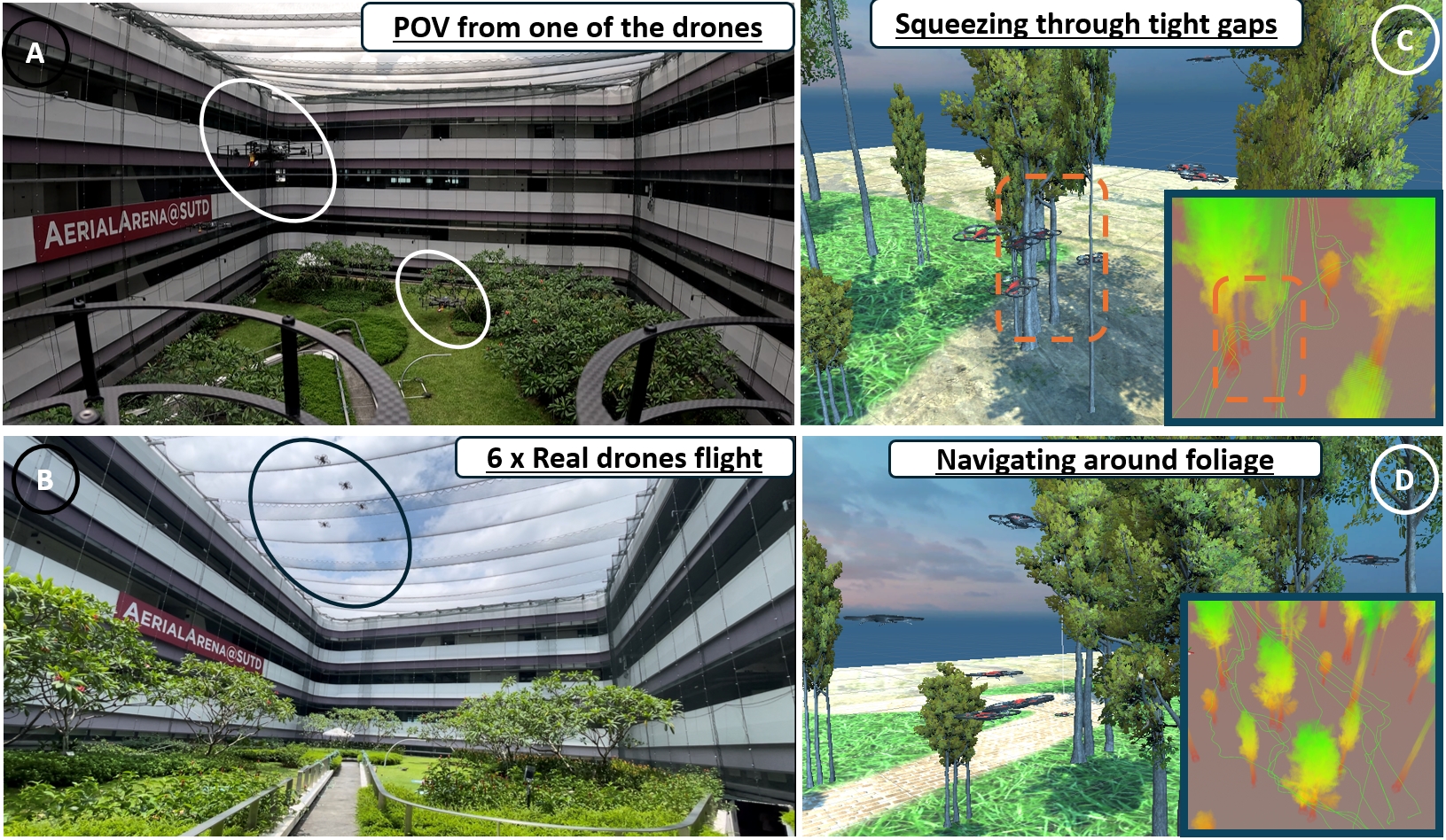}
      \caption{Hardware-in-the-loop flocking test: (A) and (B) show six physical drone flights captured from the drone's perspective and a bottom-up view. (C) and (D) show the corresponding simulated cluttered environment where the drones navigate through tight gaps and around foliage using depth image input. }
      \label{front-page}
   \end{figure}
   
Many robotic flocking systems are derived from foundational flocking models, commonly known as Artificial Potential Field (APF) methods, such as those introduced by Reynolds and Vicsek et al. \cite{reynolds1987flocks, vicsek1995novel}. To improve real-world applicability, Olfati-Saber \cite{olfati2006flocking} introduced obstacle virtual agents, representing obstacles as single points to facilitate avoidance. While numerous studies have adopted similar frameworks, real-world flocking remains an open challenge—particularly in handling environmental complexity and local minima. Despite impressive demonstrations, such as the flocking of 14 and 30 drones in real-world environments \cite{vasarhelyi2018optimized, albani2022distributed}, existing approaches often assume idealized geometric obstacles to side-step the issue of local minima. Currently, most APF strategies for overcoming local minima \cite{mcguire2021viscoelastic} are also passive, leading to slow and inefficient progress toward goals. While alternative planning-based methods (e.g., model predictive control \cite{dave2019decentralized, morgan2016swarm}) and learning-based approaches \cite{yan2021deep} to achieve collective navigation can effectively mitigate local minima, they introduce higher computational costs either during training or test time \cite{beaver2021overview}.

To this end, we advance the field of APF methods by extending the work of \cite{olfati2006flocking} and \cite{vasarhelyi2018optimized} to achieve flocking with quicker navigation towards goal while minimizing local minima. Specifically, this paper makes the following contributions:
\begin{itemize}
    \item We present GO-Flock, a hybrid framework with a novel APF serving as the reactive Collective Navigation Module that integrates inter-agent cohesion-repulsion, waypoint attraction, and obstacle avoidance. This module operates at a lower level, leveraging information from a higher-level planning Perception Module which processes depth maps to identify the required intermediate waypoints and virtual agents.
    \item To the best of our knowledge, we are the first to demonstrate an APF framework with onboard depth sensor for flocking in realistic 3D environments. We validate our approach through simulations in realistic scenarios such as forests, and showcase a hardware-in-the-loop experiment where $6$ real drones interact with $3$ virtual drones in a simulated environment.
\end{itemize}

\section{RELATED WORKS}

\subsection{Decentralized Flocking Algorithms}
According to \cite{sar2023flocking}, collective motion can be broadly categorized into swarming and flocking, with the key distinction being that flocking requires agents to maintain a largely aligned heading. Collective motion can be achieved through optimization-based methods \cite{hu2020convergent, zhou2022swarm, hu2021decentralized}, learning-based approaches \cite{schilling2021vision, hu2020vgai}, and APFs. While millisecond-range formation planning has been demonstrated in forest environments \cite{zhou2022swarm}, optimization techniques are generally computationally demanding \cite{song2023reaching}. Learning-based methods, such as those proposed by Schilling et al. \cite{schilling2021vision} and Hu et al. \cite{hu2020vgai}, have shown promise in end-to-end mapping from image pixels to control actions. However, their scalability and performance in obstacle-rich environments have yet to be thoroughly validated. APF methods are computationally efficient and have been widely applied to achieve flocking behavior. These methods can be broadly classified into two categories: leader-follower \cite{gervasi2004coordination, carpin2002cooperative, balazs2020adaptive} and decentralized (leaderless) approaches \cite{olfati2006flocking, vasarhelyi2018optimized, amorim2021self, albani2022distributed}. Leader-follower methods rely on designating a real or virtual leader, with the follower agents aligning their heading to the leader while maintaining cohesion \cite{xiong2010survey}. However, these approaches tend to be less robust in the event of leader failure. On the contrary, decentralized flocking achieves cohesive group behavior through local interactions, without relying on any designated leader, resulting in robustness and adaptability to agent failures.

Classic examples of decentralized flocking include the seminal work by Olfati-Saber \cite{olfati2006flocking}, who extended flocking principles to practical robotic systems . His work introduced a systematic framework for flocking control and the novel concept of obstacle virtual agents to facilitate static obstacle avoidance. Building on this foundation, Vásárhelyi et al. \cite{vasarhelyi2018optimized} developed a scalable and stable flocking model optimized using an evolutionary algorithm. His research culminated in a large-scale real-world flight demonstration involving a swarm of 30 drones, showcasing the feasibility of decentralized flocking in physical environments . Amorim et al. \cite{amorim2021self} pushed the boundaries of communication-less flocking by developing a proximal control algorithm based on relative position sensing, successfully demonstrating it in GNSS-denied environments. Albani el al. \cite{albani2022distributed} extended this work with self-tuning algorithms that balance group cohesion and individual navigation goals, while Horyna et al. \cite{horyna2024fast} improved state estimation accuracy with a multi-agent algorithm requiring minimal communication. 

\subsection{Local Minima, Obstacle Avoidance and Environment Complexity}
APF methods are well-known for their susceptibility to local minima, which occur when an agent becomes trapped in a region of low potential surrounded by higher potential areas. A common source of local minima is the presence of obstacles. Many existing approaches either neglect obstacles altogether \cite{albani2022distributed, balazs2020adaptive, amorim2021self, verdoucq2023bio}, assume global knowledge of the environment \cite{barnes2006swarm, barnes2007unmanned,chiun2024star}, or rely on passive strategies to escape local minima \cite{mcguire2021viscoelastic, hettiarachchi2005distributed}. Map-dependent methods are often impractical in real-world settings, while passive strategies typically require agents to become stuck, recognize their situation, and then adjust their potential function to escape, which can result in slow overall progress towards the goal. Furthermore, the success of such approaches frequently relies on neighboring agents remaining mobile to perturb the potential field and assist the trapped agent \cite{mcguire2021viscoelastic}. In single-agent navigation using APFs, local minima are also often addressed through passive methods such as simulated annealing \cite{he2020dynamic, zhu2006robot}, which introduces random uphill moves (towards higher potential) to enable escape. However, these methods are typically demonstrated only in simplified 2D environments with basic obstacle configurations, limiting their applicability to realistic 3D settings.

In fact, the application of APF in realistic 3D environments remains underexplored. Most existing works are demonstrated primarily in idealized settings with simplified obstacle representations \cite{olfati2006flocking, vasarhelyi2018optimized}, thus requiring the need for known and pre-built maps for obstacle avoidance. This reliance limits their applicability to real-world environments. An extension to Olfati-Saber’s work by Li et al. \cite{li2015flocking} tackled obstacles with arbitrary shapes by discretizing obstacle surfaces into virtual agents represented as discrete nodes. However, this approach becomes computationally expensive as obstacle complexity and scale increase, posing further challenges for real-time implementations in realistic environments.

\section{APPROACH}

\begin{table}[h]
\caption{Comparison of representative works across paradigms (O: Optimization, L: Learning, A: APFs). MC: Minimal communication, EC: Environment complexity, UE: Unknown environment, LM: Addresses Local minima.}
\label{table_example}
\begin{center}
\begin{tabular}{|p{2.8cm}|p{0.8cm}|p{0.8cm}|p{0.8cm}|p{0.8cm}|}
\hline
Name (Paradigm) & MC & EC & UE & LM \\
\hline
Zhou et al. \cite{zhou2022swarm} (O) & \xmark & \cmark & \cmark & \cmark  \\ 
\hline
Schilling et al. \cite{schilling2021vision} (L) & \cmark & \xmark & \xmark & \cmark \\
\hline
Vásárhelyi et. al \cite{vasarhelyi2018optimized} (A) & \cmark & \xmark & \xmark & \xmark \\
\hline
Maguire et al. \cite{mcguire2021viscoelastic} (A) & \cmark & \xmark & \cmark & \cmark  \\
\hline
{\color{black}\textbf{GO-Flock (A)}} & {\color{black} \cmark} & {\color{black} \cmark} & {\color{black} \cmark} & {\color{black} \cmark} \\
\hline
\end{tabular}
\end{center}
\end{table}

Our work extends the decentralized flocking literature by introducing GO-Flock, a hybrid framework that mitigates local minima through an integrated planning Perception Module. This module leverages depth maps to identify waypoints and virtual obstacle agents, which inform the downstream Collective Navigation Module governed by our novel APF formulation. The navigation module models potential functions between these point-mass agents to maintain safe distancing and ensure cohesive group movement. Beaver \cite{beaver2021overview} categorizes flocking algorithms into reactive and planning approaches and states that planning approaches, though computationally expensive, are better able to prevent local minima. Hence, our approach can be seen as a hybrid model because it combines both planning and reactive approaches. The addition of the perception module can be seen as enhancing the reactive APF methods with some degree of planning; however, unlike most planning approaches \cite{hu2020convergent, zhou2022swarm, hu2021decentralized}, our method focuses on finding a single target waypoint rather than a trajectory. Also, contrasting with prior APF methods that uses passive methods to address local minima \cite{barnes2006swarm, barnes2007unmanned, mcguire2021viscoelastic, hettiarachchi2005distributed}, our approach operates without prior map knowledge and proactively resolves local minima by selecting intermediate waypoints to guide agents around obstacles. Together with our APF formulation, Go-Flock enables robust flocking in realistic 3D environments. Our work is closely related to Olcay et al.'s framework \cite{olcay2020collective}, which integrates a single-agent navigation policy with a multi-agent collective navigation algorithm. However, Olcay et al.'s wall-following approach requires agents to follow the obstacle surface tangentially and thus has only been demonstrated to work with 2D planar surfaces.

Table \ref{table_example} compares representative works. Our approach actively resolves APF local minima and is demonstrated in a realistic 3D environment using depth sensors with minimal communication (pose only) without a pre-built map.

For clarity, we adhere to standard notation throughout this paper unless stated otherwise. The relative position vector from agent $j$ to agent $i$ is denoted as $\mathbf{p}_{i-j}$, with $\|\mathbf{p}_{i-j}\|$ representing its Euclidean distance. The velocity vector of Agent $i$ is represented by $\mathbf{v}_i$. Unit vectors are denoted with a hat, e.g., $\mathbf{\hat{v}}_i$. APF methods represent agent interactions as potential fields, with the overall potential modeled as a linear combination of pairwise inter-agent potential functions. Therefore, for better readability, we refer to target waypoint and virtual agents from $w_1$ to $w_4$, target goal as $g$, while Agent $i$ represents the ego-agent. Greek letters denote parameters and are explained in the Appendix. Refer to Fig. \ref{fig:algo_diagram} for the remaining of this section.

\subsection{Perception Module}\label{subsection: perception}

\begin{figure}[thpb]
      \centering
      \includegraphics[clip,  bb= 0 0 1300 480,  width=1.68\columnwidth]{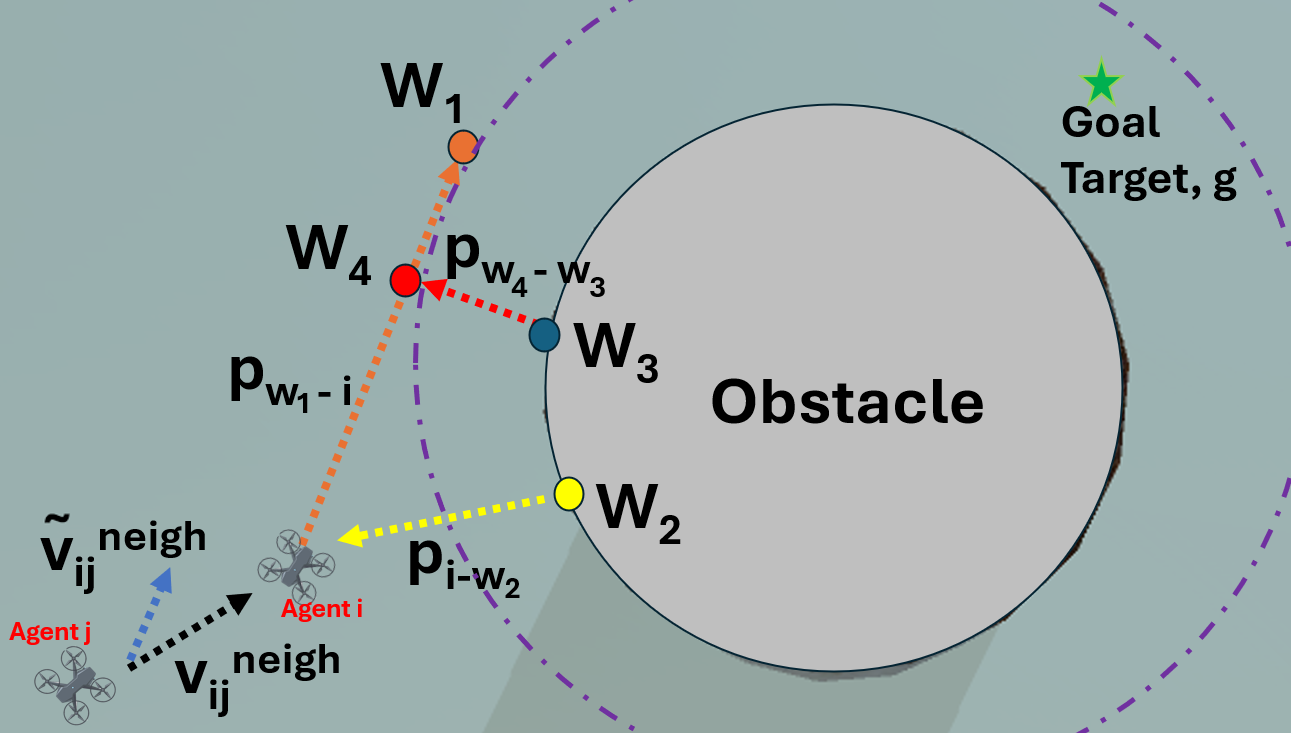}
      \caption{The color-filled circles represent the waypoint and virtual agents \(w_1\) to \(w_4\). The dotted arrows represent the potential vectors utilized from Eqn. \ref{eqn:neighbours} to \ref{eqn:final_eqn}. The dotted purple circle concentric to the obstacle represents the inflation to account for the drone's volume and safety. $w_1$ is thus chosen to be the tangent to the inflated obstacle, which minimizes deviation from the direct path to $g$. Note that \(\textbf{v}_{ij}^{neigh}\) if applied naively to Eqn. 1, would have result in \(v_{i}^{des}\) having vector components that drive Agent i towards the obstacle. To improve safety, Eqn \ref{eqn:final_eqn} and hence, \(\tilde{\textbf{v}}_{ij}^{neigh}\),  sought to remove such effects. Note that \(\tilde{\textbf{v}}_{ij}^{neigh}\) would eventually lie on the left half plane of \(\textbf{p}_{w_1 - i}\) } 
      \label{fig:algo_diagram}
   \end{figure}

Our waypoint identification method is inspired by pigeons' goal-directed navigation strategies, which balance safety and efficiency during flight \cite{lin2014through}. As shown in Fig. \ref{fig:algo_diagram}, when the direct path to the goal \( g \) is blocked, the perception module selects an intermediate waypoint \( w_1 \) that minimizes deviation from the direct path, allowing the agent to bypass the obstacle with minimal extra travel. This deviation-minimizing approach helps maintain flock cohesion over time, even though it follows a single-agent greedy strategy that may temporarily cause local splits in the flock, as discussed later in Section \ref{subsection:baseline}. In addition to waypoints, the module also identifies virtual agent positions (\( w_2 \) to \( w_4 \)) to aid in obstacle avoidance.

Several possible approaches exist for identifying waypoints and virtual agents $w_1$ to $w_4$. In this paper, we opted to convert depth maps into occupancy grid maps with an inflation margin $\delta$ to account for the drone’s volume and safety. The path generated by a 3D A* planner is used to select the target waypoint $w_1$, which is chosen near the point tangent to the obstacle’s edge with sufficient clearance to minimize bypass effort. The tangent path to the obstacle provides a locally shortest path that deviates least from the direct path to goal \cite{liu1994computation}. For obstacle avoidance, virtual agents are defined as follows: $w_2$, the nearest point on the obstacle to Agent \(i\); $w_3$, the point on the obstacle closest to \(\textbf{p}_{w_1 - i}\); and $w_4$, the nearest point on \(\textbf{p}_{w_1 - i}\) to $w_3$. While \cite{olfati2006flocking} proposes only $w_2$ for obstacle avoidance, we included $w_3$ and $w_4$. The rationale for these virtual agents will be explained in Section~\ref{section: collective nav algo}.

\subsection{Collective Navigation Module} \label{section: collective nav algo}

While the Perception Module identifies the virtual agents, the Collective Navigation Module applies potential functions between agents in order to guide a group of agents to move collectively and cohesively toward a shared goal while avoiding inter-agent collisions. Eqn. \ref{eqn:overall_eqn} shows the desired velocity command for Agent $i$, $\mathbf{v}_{i}^{des}$, which is a linear combination of the various terms. We then did a min-max scaling to scale $||\mathbf{v}_{i}^{des}||$ to within $[-\phi^{max}, \phi^{max}]$ before sending the command to the drone.

\begin{equation} \label{eqn:overall_eqn}
    \mathbf{v}_{i}^{des} = \mathbf{v}_{i}^{goal} + \mathbf{\tilde{v}}_{i}^{neigh} + \mathbf{v}_{i}^{obs} 
\end{equation}

\subsubsection{Neighbouring Agent Attraction Repulsion} \label{subsection:neighbours}

The velocity vector \( \mathbf{v}_{ij}^{neigh} \) represents the influence of neighboring agent \( j \) (\( j \in \mathcal{N}^{nbr} \)) on Agent \( i \). It is proportional to the deviation between the actual (\(|\mathbf{p}_{i-j}|\)) and desired (\(\tau\)) inter-agent distance, applied along \( \mathbf{\hat{p}}_{i-j} \). Agents move apart when too close and closer when too far. If \( \left| \tau - |\mathbf{p}_{i-j}| \right| \leq \beta \) (where \( \beta \) is a threshold), no adjustment is needed, keeping \( \mathbf{v}_{ij}^{neigh} = 0 \) for stability.

\begin{equation}\label{eqn:neighbours}
  \textbf{v}_{ij}^{neigh}=
  \begin{cases}
    \varphi_n^{ref} (\tau - |\textbf{p}_{i-j}|) \hat{\textbf{p}}_{i-j}, & \text{if $\left| \tau - |\textbf{p}_{i-j}| \right| > \beta$} \\
    0, & \text{otherwise}
  \end{cases}
\end{equation}

\subsubsection{Waypoint Attraction}
In general, this component attracts Agent \(i\) towards \(w_1\) or \(g\) depending on whether the goal is visible from Agent \(i\)'s line of sight. The upper bound of this component is the Agent-Goal Control Gain, $\varphi_g^{ref}$

\begin{equation}\label{eqn:goal}
  \textbf{v}_{i}^{goal}=
  \begin{cases}
    \min(\varphi_g^{ref} (|\textbf{p}_{w_1 - i}|), \varphi_g^{ref})  \hat{\textbf{p}}_{w_1 - i}, & \text{if \textbf{g} not visible} \\
    \min(\varphi_g^{ref} (|\textbf{p}_{g - i}|), \varphi_g^{ref})\hat{\textbf{p}}_{g - i}, & \text{otherwise}
  \end{cases}
\end{equation}

\subsubsection{Obstacle repulsion} \label{subsection:obs repulsion}
The virtual agents \(w_2\) and \(w_3\) are positioned on the obstacle surface to exert repulsive forces on Agent \(i\). The vector \(\hat{\textbf{p}}_{i-w_2}\) and \(\hat{\textbf{p}}_{w_4 - w_3}\) pushes Agent \(i\) away from the obstacle if they come within a safety distance \(\sigma^s\) (Eqn. \ref{eqn:obs_repulsion}). $\varphi_o^{ref}$ is the Agent-Obstacle Control Gain.

\begin{dmath} \label{eqn:obs_repulsion}
  \textbf{v}_{i}^{obs}=
    \varphi_o^{ref}\left( \frac{\max\left(\sigma^{s} - |\textbf{p}_{w_4 - w_3}|,0\right)}{\sigma^{s}} \hat{\textbf{p}}_{w_4 - w_3} + \frac{\max(\sigma^{s} - |\textbf{p}_{i-w_2}|,0)}{\sigma^{s}}\hat{\textbf{p}}_{i-w_2}\right)
\end{dmath}

\(\hat{\textbf{p}}_{w_4 - w_3}\) is nearly perpendicular to Agent i's direct path to \(w_1\) (\(\hat{\textbf{p}}_{w_1 - i}\)). Any velocity component aligned with \(-\hat{\textbf{p}}_{w_4 - w_3}\) could potentially move Agent \(i\) closer to the obstacle over time. More concretely, \(\textbf{v}_{ij}^{neigh}\) could produce terms that steer Agent \(i\) into obstacle. To mitigate this, we remove \(\textbf{v}_{ij}^{neigh}\)'s components that align with \(\hat{\textbf{p}}_{w_3 - w_4}\) when the agent is within a certain distance from the nearest obstacle. This adjustment ensures that there are no velocity components that will push the agent towards the nearest obstacle, enhancing overall safety. The responsibility for maintaining inter-agent collision avoidance is thereby shifted more towards the agents in safer positions.

\begin{dmath} \label{eqn:final_eqn}
    \tilde{\textbf{v}}_{ij}^{neigh} = \textbf{v}_{ij}^{neigh} - \left( \textbf{v}_{ij}^{neigh} \cdot \hat{\textbf{p}}_{w_4 - w_3} \right) \hat{\textbf{p}}_{w_4 - w_3} 
\end{dmath}

Afterwhich, \(\tilde{\textbf{v}}_{i}^{neigh}\) is simply \(\sum_{j \in \mathcal{N}^{nbr}} \tilde{\textbf{v}}_{ij}^{neigh}\)

\section{RESULTS AND ANALYSIS}

\subsection{Simulation Setup and Evaluation Metrics}

   \begin{figure}[thpb]
      \centering
      \includegraphics[clip,  bb= 0 0 750 500,  width=1.23\columnwidth]{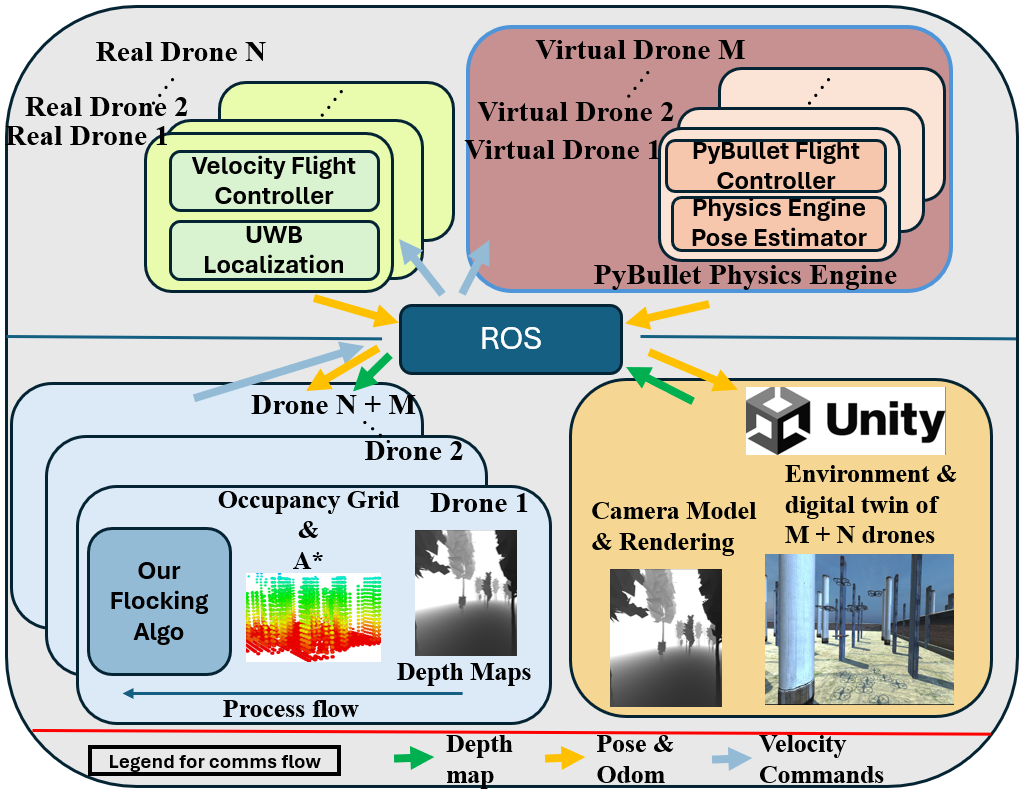}
      \caption{Hardware-in-the-loop system diagram for conducting real-world flocking experiments. Each drone's flocking algorithm is run as a separate process on the same workstation. The green, yellow and blue arrows represent communication flow. For example, velocity commands (blue arrows) are sent from the flocking algorithm and received by the real and virtual drones. }
      \label{fig:system diagram}
   \end{figure}

Fig. \ref{fig:system diagram} illustrates our system diagram for conducting Hardware-in-the-loop (HITL) experiments that integrate real and virtual drones. This hybrid setup offers experimental flexibility with semi-realistic testing environment. The architecture consists of four main modules operating in parallel: (1) a real-time PyBullet physics engine for virtual drones, adapted from gym-pybullet-drones \cite{panerati2021learning}; (2) real drones equipped with onboard flight controllers and UWB localization systems; (3) our decentralized GO-Flock; and (4) Unity3D, which provides digital twin visualization of both real and virtual drones and includes high-fidelity depth map rendering. ROS (Robot Operating System) acts as a middleware to facilitate communication between these modules. Each agent's flocking module is executed as an independent process node, enabling decentralized operation. These processes are run in parallel on a 24-core workstation, with system capacity evaluations confirming that all processes operate within the workstation's computational limits. While this distributed approach introduces latency and synchronization challenges, it enables asynchronous decision-making and decentralized control, resulting in a more realistic and robust testing environment. For full virtual experiments, the same system architecture is used by omitting the real drone module. Given that this work serves as a proof-of-concept, certain functionalities such as the removal of occluding drones from the depth image have not yet been implemented. Currently, Unity3D renders depth maps with invisible drones as a temporary solution. Future work will integrate these missing capabilities to develop a fully decentralized flocking system similar to \cite{zhou2022swarm}.

We believe that it is advantageous for the flock to occasionally split to prioritize speed and safety in dense-obstacle environments. Consequently, we assess performance based on the centroid of the drones rather than individual distances or connectivity metrics. Hence, we adapted the metrics used in \cite{verdoucq2023bio} for our purposes.

\subsubsection{Dispersion, D(t)} A cohesion metric that quantifies the spread of drones from the centroid. Low \(D(t)\) indicates insufficient separation, which can compromise safety, while excessively high values suggest undesirable dispersion. \(M\) represents total number of drones, and \(\lVert \mathbf{x}_i(t) - \mathbf{c}(t) \rVert\) denotes the Euclidean distance of each drone from the centroid.

\begin{figure*}[!t]
    \centering
    \includegraphics[width=\textwidth, height=4.5cm]{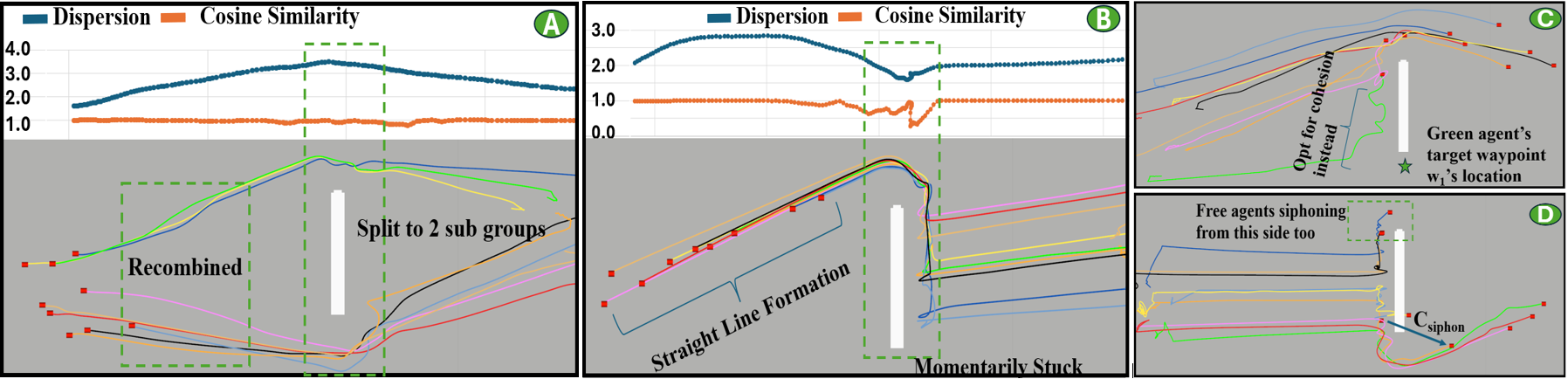}
    \caption{The white rectangle represents an obstacle in the environment. Each multi-colored path shows the past trajectory of an agent, illustrating where it has moved over time. The red dots indicate the agents' current positions at the moment captured in the snapshot. (A) A sample run using GO-Flock showing the separation of a flock into 2 subgroups as well as its corresponding $D(t)$ and $CS(t)$ metrics. (B) A typical run using Maguire et al.'s showing the siphoning effect of the algorithm and it corresponding $D(t)$ and $CS(t)$ metrics.  (C) Increasing $\varphi_n^{ref}$ and thus inter-agent cohesion factor could reduce stray agent incidents from occurring. Instead of being attracted to the target waypoint $w_1$, the green agent opted to move in cohesion with the rest of the group. (D) A sample run showing Maguire et al.'s method resulting in separation of the flock. Since $\textbf{C}_{siphon}$ (the blue arrow pointing from the stuck agent to the nearest free agent) was directed toward the obstacle—because the free agent had already moved ahead—the stuck agents ended up colliding with the obstacle.  }
    \label{fig:fullwidth_pen}
\end{figure*}

\begin{equation}\label{eqn: dispersion}
    D(t) = \frac{1}{M} \sum_{i=1}^{M} \lVert \mathbf{x}_i(t) - \mathbf{c}(t) \rVert
\end{equation}

\subsubsection{Cosine Similarity, C(t)} The degree of alignment between individual agent's velocity and the average velocity of the group. \( \mathbf{\bar{v}}(t) \) is the mean velocity of the flock at time \( t \). This value range between 0 and 1 with 1 being most aligned.

\begin{equation}
    \text{C(t)} = \frac{1}{M} \sum_{i=1}^{M} \frac{\mathbf{v}_i(t) \cdot \mathbf{\bar{v}}(t)}{\lVert \mathbf{v}_i(t) \rVert \cdot \lVert \mathbf{\bar{v}}(t) \rVert}
\end{equation}

\subsubsection{Average Speed, AV} We measure the average speed the flock needs to navigate to their common goal, $\textbf{g}$. $T$ refers to the time where all agents are at least 3m from $\textbf{g}$.

\begin{equation}\label{eqn:as}
    \text{AV} = \frac{\lVert \frac{1}{M}\left(\sum_{i=1}^{M}\mathbf{x}_i(0)\right) - \frac{1}{M}\left(\sum_{i=1}^{M}\mathbf{x}_i(T)\right) \rVert}{\text{Total Time Taken}}
\end{equation}

\begin{table}[h]
\caption{Average metrics with standard deviation comparison}
\label{table_2}
\begin{center}
\begin{tabular}{|p{0.7cm}|p{1.6 cm}|p{1.3cm}|p{1.3cm}|p{1.3cm}|}
\hline
Metric & \multicolumn{2}{|c|}{Experiment 1} & \multicolumn{2}{|c|}{Experiment 2} \\
\hline
& Maguire et al. & Ours & Baseline & Ours \\
\hline
D & \textbf{2.27 \textpm{0.14}} & 2.30 \textpm{0.37} & 3.20 \textpm{0.47} & \textbf{2.60 \textpm{0.38} } \\ 
\hline
CS & 0.78 \textpm{0.04} & \textbf{0.93 \textpm{0.03} } & 0.74 \textpm{0.10} & \textbf{0.90 \textpm{0.08} }\\
\hline
AV  & 1.17 \textpm{0.10} & \textbf{1.25 \textpm{0.06} } & 0.98 \textpm{0.12} & \textbf{3.20 \textpm{0.11} } \\
\hline
\end{tabular}
\end{center}
\end{table}

\subsection{Baseline Comparisons} \label{subsection:baseline}

In Experiment 1, we benchmark our approach against Maguire et al.'s \cite{mcguire2021viscoelastic} to evaluate strategies for overcoming local minima. To align with their original test conditions, we conducted our benchmark in an infinitely tall single-obstacle setting (5m × 0.5m). In this setup, 9 virtual drones completed 20 runs each, with varying start and end points. Maguire et al.'s approach \cite{mcguire2021viscoelastic} enhances traditional local minima-prone APF paradigms \cite{olfati2006flocking, vasarhelyi2018optimized} by incorporating a siphoning velocity component, $C_{\text{siphon}}$, which helps pull stuck agents toward free ones. Accordingly, we first adapt GO-Flock into a baseline similar to \cite{vasarhelyi2018optimized}, by removing the perception module. The baseline follows Eqs.~\ref{eqn:overall_eqn}–\ref{eqn:final_eqn}, with \(w_1 = g\) fixed and virtual agent components (\(w_3, w_4\)) removed (henceforth referred to as Baseline). Then, we implemented Algorithm~1 from Maguire et al.'s paper to determine $C_{\text{siphon}}$ and incorporated it with the Baseline to create the siphoning effect. Finally, we tuned it for reasonable performance. Our adaptation of Maguire et al.'s algorithm retains the essence of their work but yet contains APFs that are largely similar to GO-Flock for a fair comparison. We averaged \(D(t)\) and \(CS(t)\) over time and runs, and \(AV\) over runs, reporting the results in Table~\ref{table_2}.

While both approaches showed similar average dispersion, GO-Flock exhibited a higher standard deviation than Maguire et al.'s method. Maguire et al.'s approach consistently produced similar flock patterns, using free agents to extract stuck agents as a group (Fig.~\ref{fig:fullwidth_pen}B). Its dispersion briefly dipped to 1.5 at the obstacle due to local minima, before rising to nearly 3.0 as agents escaped one by one, forming a straight line formation. In contrast, GO-Flock primarily maintained cohesion as a group, though it occasionally split to navigate obstacles due to differing target waypoint \( w_1 \) (Fig.~\ref{fig:fullwidth_pen}A). Although GO-Flock's maximum dispersion \( D(t) \) of 3.5 m, occurring just before obstacle traversal, is only slightly higher than Maguire et al.'s 3.0 m, its deviation-minimizing principle in selecting \( w_1 \) likely correlates the maximum \( D(t) \) with the size of the obstacle. Consequently, GO-Flock's average dispersion depends on the obstacles and environment layout. To enhance GO-Flock's dispersion performance, inter-agent cohesion can be strengthened by adjusting $\varphi_n^{ref}$, minimizing occurrences of stray agents during navigation. As shown in Fig.~\ref{fig:fullwidth_pen}C, a stray agent (green path) was attracted back to the group, prioritizing cohesion over navigation. Finally, while Maguire et al.'s approach generally resulted in largely similar flock trajectories, it could also lead to a split in the flock, as shown in Fig.~\ref{fig:fullwidth_pen}D, although this was a rare occurrence. The split occurred when two separate agents tries to siphon out stuck agents from two different locations. This means that the performance of Maguire et al.'s approach also largely depended on the topology of the flock in relation to the obstacle.

In the context of CS and AV, GO-Flock outperforms Maguire et al.'s approach primarily because Maguire et al.'s method is passive. It requires the agent to first become trapped in a local minimum, depending on a neighboring free agent to help guide it out. This leads to low $CS(t)$ values as the agents approach the obstacle and enter the local minima, where a minimum $CS(t)$ value of 0.45 was recorded (Fig. \ref{fig:fullwidth_pen}B). In contrast, GO-Flock employs an active strategy to prevent the agent from becoming stuck in local minima, resulting in an average $CS$ metric of approximately 0.93. 

We further evaluate GO-Flock's performance in a large scale, obstacle dense environment. As Maguire et al.'s work relies on a proxy signal (e.g., goal visibility) to guide stuck agents, it is less effective in such environment. Hence, in Experiment 2, we compare GO-flock against Baseline by running simulations featuring 9 virtual drones navigating in an environment with randomly placed and infinitely tall obstacles (2m diagonal, ~3m apart) at \(\phi^{max} = 2\)~m/s. Both frameworks were tested under same parameters (Appendix) across 30 runs and their average metrics reported in Table~\ref{table_2}.  
 
GO-Flock outperforms Baseline in all 3 metrics, significantly reducing the flocking time to goal without compromising cohesion or velocity alignment. Although waypoints may cause temporary splits as agents maneuver around obstacles, we observed that agents consistently sought to regroup whenever possible. This behavior arises because the virtual waypoint, $w_1$, is selected based on the principle of minimizing deviation from the direct path. Over-fragmentation into smaller subgroups was rare. In addition, combining this with cohesion effects described in Eqn. \ref{eqn:neighbours} further ensures that agents remain closely grouped over time. In contrast, Baseline struggled in dense obstacle environments, frequently causing agents to become stuck and fragment into multiple subgroups, especially when navigating planar surfaces like cuboids. Increasing inter-agent cohesion—by adjusting gains and \( \left| \mathcal{N}^{nbr} \right| \)—reduced dispersion to \(2.94 \pm 0.62\), but at the cost of longer travel times (AV of \(0.55 \pm 0.14\)). Finally, GO-Flock achieved a significantly higher CS metric, averaging 0.90 compared to the baseline's 0.74. The lower CS for Baseline was due to agents frequently getting stuck at obstacles.

\subsection{Obstacle Avoidance Ablation Study}\label{section:ablation}

This section aims to justify the selection of virtual agents \(w_3\), and \(w_4\), as well as Eqn. \ref{eqn:final_eqn} for obstacle avoidance. The role of virtual agents \(w_3\) and \(w_4\) becomes particularly important when the flock split to circumnavigate obstacles. Without these agents, the cohesion forces described in Eqn. \ref{eqn:neighbours} may cause the separated groups to pull toward one another, drawing them closer to the obstacle.

We performed an ablation study in a simplified environment featuring a single infintely-tall obstacle with a diagonal of 2.5m and evaluate 6 drones' ability to navigate without collisions. We set \( \left| \mathcal{N}^{nbr} \right| \) to be 3, and the reference radius \(\tau\) was fixed at 2.0 m. These parameters were selected to ensure that, when the flock splits, the attraction forces of the group traversing the opposite side of the obstacle, due to Eqn. \ref{eqn:neighbours} could still be felt. The study was carried out in three modes: (1) a control mode with obstacle repulsion disabled (i.e., deactivating Eqs. \ref{eqn:obs_repulsion} and \ref{eqn:final_eqn}), (2) using only the static obstacle repulsion component involving the virtual agent \(w_2\) and changing Eqn. 5's \(\hat{\textbf{p}}_{w_4 - w_3}\) to \(\hat{\textbf{p}}_{i - w_2}\) , and (3) using only virtual agents \(w_3\) and \(w_4\) to avoid obstacles. We ran each experimental group a total of 20 times.

\begin{figure}[thpb]
      \centering
      \includegraphics[clip,  bb= 0 0 1300 400,  width=1.37\columnwidth]{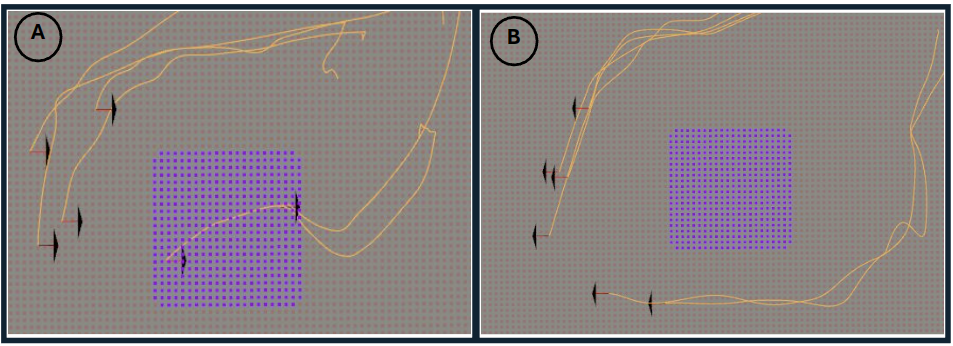}
      \caption{Orange lines show the path undertaken by the agents from Experimental Group 2 (Panel A) and 3 (Panel B). The purple square shows an infinitely tall cuboid obstacle. (A) As the agents split into 2 subgroups, the cohesive effects drew the smaller subgroup into obstacle. (B) Both subgroups were able to circumnavigate the obstacle successfully and rejoined. } 
      \label{fig:abla}
   \end{figure}

Unlike Experimental group (3), groups (1) and (2) consistently failed to navigate the environment successfully. Figures \ref{fig:abla}A and \ref{fig:abla}B depict the paths taken by agents in groups (2) and (3), respectively. The failures observed in group (1) were anticipated, as this group lacked any obstacle avoidance mechanisms. Consequently, inter-agent cohesion forces led the flock to collide with obstacles. The failure of group (2), however, can be attributed to the execution of Eqn. \ref{eqn:final_eqn} using only the influence of \(w_2\). This resulted in \(\tilde{\textbf{v}}_{ij}^{neigh}\) containing components that deviates from \(\textbf{p}_{w_1 - i}\) and could direct the drone towards obstacles due to inter-agent cohesion effects, thus increasing the likelihood of collision. In contrast, when both \(w_3\) and \(w_4\) were incorporated in Eqn. \ref{eqn:final_eqn}, the resulting \(\tilde{\textbf{v}}_{ij}^{neigh}\) aligned more closely with \(\textbf{p}_{w_1 - i}\), effectively preventing the agent from approaching obstacles. There were no inter-agent collisions too. Furthermore, we assessed obstacle avoidance performance by applying \(w_3\) and \(w_4\) in Eqn. \ref{eqn:final_eqn} without incorporating Eqn. \ref{eqn:obs_repulsion}. This approach demonstrated similar performance. However, for enhanced safety and robustness, we retained Eqn. \ref{eqn:obs_repulsion} within our framework to account for potential unforeseen scenarios.

\subsection{Test with Realistic Environments}

\begin{figure*}[!t]
    \centering
    \includegraphics[width=\textwidth]{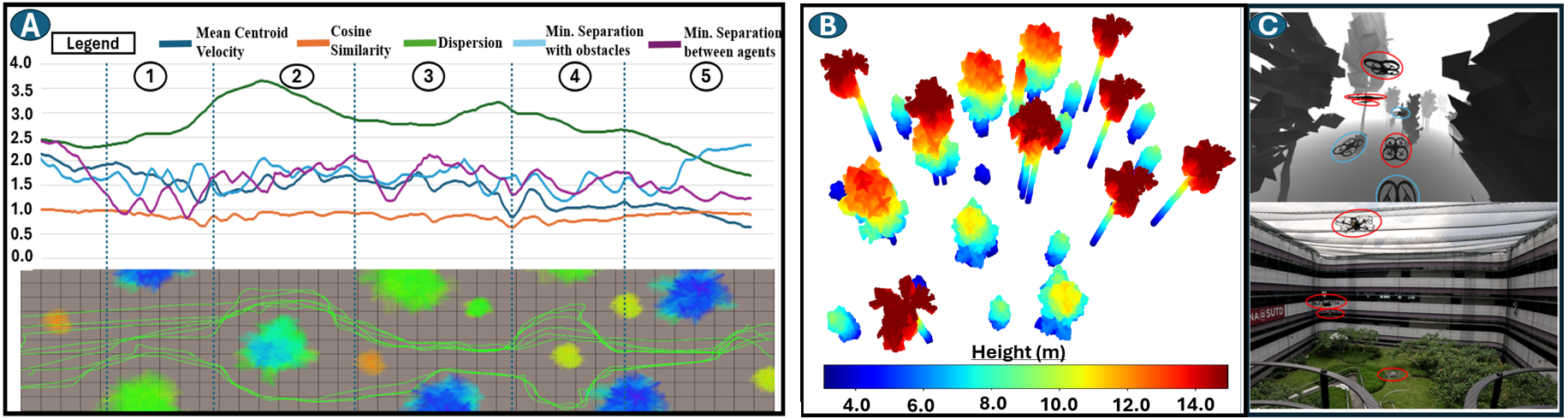}
    \caption{(A) A sample forest simulation run with the path trajectory and metrics. (B) shows an example of the forest built for the test with realistic environments. (C) The digital twin representation mirrors the exact movements of the six physical drones, along with three additional virtual drones for experimental convenience, operating in a simulated Unity3D forest. (Red represents real drones; Blue represents virtual drones and thus would be invisible in the image below)  } 
    \label{fig:fullwidth_image}
\end{figure*}

We test the effectiveness of GO-Flock with 9 drones navigating through a forest environment with varied tree and bush sizes. We randomized two forest environments by varying tree placement and size and performed 30 runs for each environment (See Fig. \ref{fig:fullwidth_image}B). For each run, the starting and end location are randomized. The forest environments had an approximate canopy coverage of 25\% around a 30m x 40m area. Most traversable gaps between trees range from between 2.5m to 5m. As the drones navigate, they typically fly between an altitude of 3m to 12m. The drones were tested at two speed ranges: low ($\phi^{max} = 1.0$ m\(/\)s) and high ($\phi^{max} = 2.0$ m\(/\)s). A run was considered failed if a collision occurred, defined as an object coming within 30 cm of the drone's center of mass. Both speed groups achieved similar success rates of approximately 90\%. Failures were primarily attributed to two factors: (1) inaccurate mapping, which affected the estimation of virtual agent positions, particularly near obstacles, and (2) agents attempting to fly through narrow gaps of less than 2m, while \(\sigma^s\), the safety distance parameter, was set to 1.5, causing oscillatory motion when navigating these tight spaces. For successful runs, the average minimum inter-agent and agent-obstacle distances per run ranged from 0.9 to 1.1m. Fig. \ref{fig:fullwidth_image}A presents the path and time-plots of key metrics for a sample run. In general, we noted that the inter-agent and agent-obstacle separation distance remains healthy throughout navigation. The agents were also generally headed in the same direction with a high CS metric. The dispersion only momentarily increases when circumnavigating obstacles but quickly recombined upon opportunity.

While reviewing the paths taken by the agents during the runs, we observed some notable behaviors. Although navigating through tight gaps led to some failures, we also recorded successful passages through gaps as narrow as 2 meters, as illustrated in Fig. \ref{front-page}C. For example, a group of 3 to 4 drones successfully squeezed through a narrow gap between two trees without collisions. Additionally, Fig. \ref{front-page}D shows instances where the drones adeptly avoided the foliage, with some opting to fly above or below the foliage.

We further validate the practical applicability of our framework in semi real-world scenarios by conducting Hardware-in-the-Loop (HITL) experiments with our architecture depicted in Fig.~\ref{fig:system diagram}. All computations, including those for real drones, were handled by a ground workstation. Localization was achieved using Ultra-Wideband (UWB) technology, which provided 0.5-meter accuracy across all three axes. Commands were sent to the real drones, and their pose and odometry data were received at a rate of 10~Hz. Although we observed a slight velocity tracking error of approximately 0.3~m/s for the real drones, we noted similar performance to that of the fully virtual drone setup. The dispersion and cosine similarity metric remained reasonable throughout the flight. Additionally, the minimum inter-agent and obstacle-agent distances were consistently conservative, with separations of over 1 meter. These results show that our algorithm remains robust under limited real-world conditions, effectively managing collision avoidance and maintaining cohesive flock behavior despite realistic noise and dynamic environmental factors.

\subsection{Computational Efficiency}

APF-based methods are computationally efficient. Even with the perception module (Section~\ref{subsection: perception}), GO-Flock remains lightweight. To demonstrate this, we implemented it on a Radxa Zero, a quad-core ARM Cortex-A53 single-board computer. The Perception Module ran at 5 Hz, while the Collective Navigation Module operated at 30 Hz. Over a one-minute test, the average CPU load was $75\%$, well within the processor’s capacity.

\section{CONCLUSIONS}

In this paper, we propose a novel hybrid framework for flocking that integrates reactive APFs with a planning-based perception module. This integration mitigates local minima issues, enhances obstacle avoidance, and preserves conventional flocking behaviors. We compare GO-Flock against baseline methods and demonstrate that, despite generating waypoints $w_1$ using a greedy single-agent approach, the principle of least deviation from the direct path enables the flock to remain largely cohesive. Furthermore, we show that the novel APF in the collective navigation module is essential for the effectiveness of this hybrid framework. We further validated GO-Flock's performance in forest environments through simulations as well as through HITL experiments.

As a proof of concept, some features, such as drone removal, remain unimplemented, with Unity3D using depth maps with invisible drones as a workaround. Additionally, HITL experiments offer limited real-world validation of GO-Flock. Future work will integrate all components, including drone masking, and deploy the system in open environments with stronger environmental factors like wind.


\section*{APPENDIX}

\begin{table}[h]
    \centering
    \renewcommand{\arraystretch}{1.3} 
    \begin{tabular}{|l|l|l|}
        \hline
        \textbf{Parameter} & \textbf{Description} & \textbf{Values} \\
        \hline
        $\varphi_n^{ref}$  & Inter-agent Control Gain  & 6 \\
        $\varphi_g^{ref}$  & Agent-goal Control Gain  & 6 \\
        $\varphi_o^{ref}$  & Agent-Obstacle Control Gain  & 12 \\
        $\tau$  & Inter-agent Desired Distance & 3 \\
        $\sigma^s$  & Agent-Obstacle Safety Distance  & 1.5 \\
        $ \left| \mathcal{N}^{nbr} \right|$  & Target Control Gain  & 3 \\
        $\delta$  & Occupancy Grid Obstacle Inflation  & 0.5 \\
        $\phi^{max}$ & Max Flight Speed  & 2 \\
        \hline
    \end{tabular}
    \caption{Parameter values used in GO-Flock for Experiments}
    \label{tab:parameters}
\end{table}

\bibliographystyle{IEEEtran}
\bibliography{root}

\end{document}